\documentclass[11pt]{article}
\usepackage{acl}

\usepackage{times}
\usepackage{latexsym}
\usepackage[T1]{fontenc}
\usepackage[utf8]{inputenc}
\usepackage{microtype}
\usepackage{inconsolata}
\usepackage{graphicx}

\title{Before the Labels: How Dataset Construction Shapes Suicidality Detection in Clinical Text}

\author{
Priyanshi Garg$^{\dagger}$\thanks{Equal contribution.}, Ishita Rao$^{\dagger}$\footnotemark[1], Jieqiong Ding$^{\ddagger}$\footnotemark[1], Amandalynne Paullada$^\dagger$\\
$^\dagger$Department of Linguistics, University of Washington, Seattle, WA, USA\\
$^\ddagger$Human Centered Design \& Engineering, University of Washington, Seattle, WA, USA\\
\texttt{\{pgarg2, ishirao, jding07, paullada\}@uw.edu}
}

\begin{document}

\maketitle
\begin{abstract}
Clinical NLP increasingly relies on electronic health record (EHR) data
to detect suicidal behaviors, treating clinical documentation as more
reliable ground truth than social media. We argue that this framing
obscures how EHR-based suicidality datasets encode a particular
operationalization of suicidality, shaped by who authors the data,
how episodes are bounded, and how ambiguity is resolved. We ground
this argument in a case study of the ScAN dataset \cite{Rawat_2022},
built over MIMIC-III clinical notes. We show how governance
constraints, ICD-based cohort selection, single-annotator labeling,
and hospital-stay-level aggregation produce labels that reflect clinician-documented judgments, treat suicidality as a bounded episode, and
assume that intent can be reliably inferred from documentation. A linguistic analysis demonstrates that identical labels subsume heterogeneous clinical framings differing in temporality, negation, and uncertainty. We argue that clinical NLP should examine the assumptions embedded in suicidality datasets before interpreting their labels as ground truth. 
\end{abstract}

\section{Introduction}

Suicide is a leading cause of death worldwide, and detecting suicidal
behavior from text is a high-stakes task in clinical NLP. Recent work has increasingly turned to electronic health records (EHRs) as a data source, motivated by well-documented concerns with social media data around consent, self-report reliability, and generalization \cite{chancellor2019, ernala2019methodological, harrigian2020state}. EHRs offer structured clinical documentation, multi-source corroboration, and longitudinal patient histories.

However, systematic reviews have found that predictive models for suicide perform only marginally better than chance across a range of clinical settings \cite{belsher2019prediction, franklin2017risk, large2016metaanalysis}. This poor predictive performance warrants scrutiny of, among other factors, the training data and the assumptions encoded within the labels. EHRs should not be treated as direct access to suicidality itself.
Clinical notes are documentation-mediated records: they synthesize patient
report, clinician observation, prior records, collateral information,
institutional requirements, and risk-management practices. This mediation
is especially important for suicidality, where disclosure and documentation
depend on the clinical interaction in which risk is assessed. Prior work
shows that healthcare professionals vary in how they ask about suicide
risk, including question designs that may invite denial or reduce
elaboration \cite{mccabeHowHealthcareProfessionals2017}. Mental health
professionals also vary in how consistently they assess suicidal thoughts
and behaviors, with assessment practices shaped in part by comfort working
with suicidal individuals \cite{roushMentalHealthProfessionals2018}. Thus,
the absence, presence, or wording of suicidality evidence in EHR text
reflects both clinical phenomena and the conditions under which those
phenomena become documented. 

Prior work has also examined ethical tensions in mental health inference,
demographic bias in diagnosis and clinical documentation, and documentation
quality in clinical NLP. For example, racial disparities in psychiatric
diagnosis can shape what enters the clinical record
\citep{schwartzRacialDisparitiesPsychotic2014}, and stigmatizing language
in EHR notes has been shown to vary across patient groups
\citep{sun2022negative}. These concerns are often treated separately. We
propose \emph{operationalization} as a unifying lens: dataset construction
decisions collectively determine what counts as evidence, how it is
categorized, what distinctions are preserved, and what forms of uncertainty
are collapsed.





Our position is that EHR-based suicidality datasets encode a specific
operationalization of suicidality:

\begin{quote}
\textit{Documentation-mediated}, reflecting suicidality as represented in clinician-authored records rather than as an unmediated patient state;
\textit{episodic}, bounding suicidality to discrete hospital stays
rather than modeling it as a longitudinal process; and
\textit{intent-resolved}, imposing binary distinctions between
suicidal and non-suicidal self-harm even when the clinical text is
ambiguous.
\end{quote}

This is not an argument that clinical documentation is unreliable, nor that patient self-report is a bias-free alternative. Suicidality assessment is
clinically difficult, relational, and uncertain. Rather, our claim is that
when clinical documentation is converted into NLP labels, those labels
should be interpreted as structured operationalizations of documented
clinical evidence, not as neutral ground truth. 

We ground this argument in a case study of the ScAN dataset
\cite{Rawat_2022}, examining how access constraints, cohort design,
and annotation decisions shape the labels and how those choices appear
in the linguistic structure of annotated spans.

Our contributions are:
\begin{itemize}
    \item A position characterizing EHR-based suicidality datasets as
    encoding a documentation-mediated, episodic, and intent-resolved
    operationalization of suicidality
    \item A case study of ScAN tracing how governance, data sourcing,
    and annotation design give rise to this operationalization
    \item Empirical evidence that these construction choices are
    associated with systematic variation in linguistic framing and
    labeling patterns
\end{itemize}

\section{ScAN as a Case Study}

ScAN \cite{Rawat_2022} is an annotation layer for suicide attempts
(SA) and suicidal ideation (SI) built over a subset of MIMIC-III, a
de-identified clinical electronic health record (EHR) dataset from
Beth Israel Deaconess Medical Center in Boston \cite{Johnson_2016}.
It contains 12,759 clinical notes across 697 hospital stays from 669
patients, with 19,690 span-level annotations. SA labels distinguish
between positive, negative, unsure, and neutral cases based on inferred
intent; SI labels capture presence or absence of ideation. We use ScAN
as a case study to examine how EHR-based suicidality labels are shaped
by the documentation, cohort, and annotation choices through which
clinical text becomes machine-learning data.

Clinical notes often
synthesize patient report, clinician observation, collateral information,
prior records, institutional requirements, and risk-management practices.
We use \emph{documentation-mediated} to mean that the resulting NLP labels
are labels over documented clinical evidence, not direct measurements of
suicidality itself. Below, we trace three dimensions of this
operationalization: documentation-mediated perspective, episodic
temporal framing, and intent resolution.

\subsection{Documentation-Mediated: Governance, Access, and Perspective}

ScAN inherits the governance model of MIMIC-III. Although annotations
and code are publicly released, the underlying clinical text requires
credentialing through PhysioNet, completion of human subjects training,
and agreement to a Data Use Agreement. Reproducibility is therefore
conditional on institutional resources and compliance capacity.

The source text is clinician-authored documentation. This does not mean
that patient perspectives are absent from the notes: patient reports,
statements, and histories may be represented within clinician-authored
text. However, direct patient-authored self-report, intake questionnaires,
and patient-facing communications are outside the annotation scope. The
dataset therefore represents suicidality as it appears in clinical
documentation, rather than as it might appear across all possible sources
of patient expression.

The annotation process further mediates the relationship between patient
experience and dataset labels. Only notes from 24 pre-selected clinical
section types were retained for annotation, based on expected relevance
to suicidality. Annotation was performed by a single trained annotator
supervised by a physician, who independently reviewed 330 of 12,759 notes
(2.6\%). Disagreements were resolved by updating labels to match the
physician review. Thus, the labels reflect the ScAN annotation protocol
as applied to clinician-authored documentation under partial physician
review. They should be interpreted as structured judgments about
documented evidence, not as neutral or exhaustive records of suicidality.

De-identification procedures, including date shifting and age masking
for patients over 89, further mediate the relationship between the data
and the clinical events they represent \cite{Johnson_2016}.

\subsection{Episodic: Cohort Selection and Temporal Flattening}

ScAN's cohort is seeded using ICD codes associated with suicide and
overdose, restricting the dataset to hospital stays where suicidality
or self-harm was already suspected, documented, or coded. ICD codes
capture only a fraction of suicidality-related events, approximately
3\% of SI and 19\% of SA by some estimates \cite{Anderson_2015}, so
patients whose suicidal behavior went uncoded are outside the dataset.

Labeling is performed at the hospital-stay level. This design is
practical for constructing an EHR-based dataset, but it also frames
suicidality as a bounded hospital episode. Temporal distinctions, such
as whether an event occurred during the current admission, in a recent
prior encounter, or years earlier, are not directly encoded in the label
schema even when such distinctions appear in the clinical text. This
matters because suicidality is often longitudinal and recurrent rather
than confined to a single encounter. Many individuals who die by suicide
sought care well before their deaths \cite{Kessler_2020}, and ideation
may recur across encounters.

\subsection{Intent-Resolved: Annotation Design and Category Collapse}

ScAN's annotation scheme uses CDC-based definitions that distinguish
suicidal from non-suicidal self-harm, requiring intent to be inferred
from clinical documentation. Non-suicidal self-injury (NSSI) is excluded
by design.

Cases where intent is unclear are labeled ``unsure,'' but this category
is collapsed with ``negative'' in the downstream modeling pipeline used
by the companion ScANER system. This merges qualitatively different
situations: documentation that explicitly negates suicidal intent,
documentation where intent is not specified, and documentation where the
available evidence is genuinely ambiguous. As we show in
Section~\ref{sec:framing}, these categories have distinct linguistic
profiles. The issue is not that the original annotation decision was
unreasonable, but that downstream label collapse can erase clinically
meaningful uncertainty.

\subsection{Dataset Composition}
\label{sec:demographics}

ScAN's single-site origin produces a demographically narrow dataset:
70\% White, 92\% English-speaking, and concentrated in the 25--54 age
range, which includes approximately 70\% of patients. No non-White
racial or ethnic group exceeds 22 admissions, making subgroup analysis
infeasible for most categories. The structured demographic data also
contains some ambiguities; full details and tables are in
Appendix~\ref{app:demographics}. We therefore treat demographic
composition primarily as a description of dataset scope, rather than as
the basis for subgroup claims.

\subsection{Linguistic Framing Analysis}
\label{sec:framing}

If labels flatten temporal, epistemic, and intentional distinctions
present in the clinical text, then spans sharing the same label should
exhibit heterogeneous linguistic profiles. We test this expectation
through a focused framing analysis.

We analyze 15,585 annotated spans after excluding those with missing
metadata. For each span, we identify negation, historical reference,
and uncertainty using lexical indicators and MedSpaCy's ConText
algorithm \citep{harkema2009context}. We classify a span as
\emph{unmodified} when both methods agree on the absence of all three
modifier types. This is a descriptive linguistic category, not a
clinical judgment: ``unmodified'' means that our detectors did not find
negation, historical reference, or uncertainty governing the span. It
does not imply clinically active suicidal intent. Details and
inter-method agreement rates are in Appendices~\ref{app:framing-method}
and~\ref{app:signal-agreement}.

\paragraph{Within-label heterogeneity.}
Spans sharing the same label exhibit substantial variation in framing
(Table~\ref{tab:framing-by-type}). Two-thirds of SA spans are
unmodified, but only about half of present-SI spans are. Of particular
note, 27.8\% of present-SI spans contain historical markers, indicating
that ideation labeled as present may still be documented in relation to
prior psychiatric history rather than as a purely present-tense event.

\begin{table}[t]
\centering
\small
\caption{Framing profiles by span type. ``Unmodified'' means no detected
negation, historical reference, or uncertainty marker.}
\label{tab:framing-by-type}
\begin{tabular}{@{}lrrrr@{}}
\hline
\textbf{Span type} & \textbf{$n$} & \textbf{\% Unmod.} & \textbf{\% Hist.} & \textbf{\% Neg.} \\
\hline
All SA        & 14,048 & 66.8 & 13.2 &  5.8 \\
Present-SI    &    897 & 52.2 & 27.8 &  7.0 \\
Absent-SI     &    554 & 13.5 & 16.1 & 65.2 \\
\hline
\end{tabular}
\end{table}

\paragraph{Ambiguous categories and the cost of collapse.}
The categories that encode uncertainty show the clearest effect
(Table~\ref{tab:framing-ambiguous}). ``Unsure'' spans are frequently
unmodified but contain elevated uncertainty markers (17.1\%), reflecting
epistemic ambiguity in the documentation. Unspecified-intent spans
(T14.91) are predominantly historical (67.8\%), indicating that lack of
intent specification often co-occurs with temporal displacement. When
these categories are merged with ``negative'' labels in downstream
pipelines, epistemic uncertainty and temporal displacement are collapsed
into a single class.

\begin{table}[t]
\centering
\small
\caption{Framing profiles of selected SA categories.}
\label{tab:framing-ambiguous}
\resizebox{\columnwidth}{!}{%
\begin{tabular}{@{}lrrrrr@{}}
\hline
\textbf{Category} & \textbf{$n$} & \textbf{\% Unmod.} & \textbf{\% Hist.} & \textbf{\% Neg.} & \textbf{\% Unc.} \\
\hline
Intentional (X71--X83) & 1,271 & 83.3 & 12.1 & 1.1 & 1.7 \\
Unsure                 & 2,237 & 61.6 & 19.3 & 6.8 & 17.1 \\
Unspecified (T14.91)   &   770 & 19.0 & 67.8 & 3.4 & 5.1 \\
\hline
\end{tabular}}
\end{table}

\paragraph{Variation across documentation contexts.}
Framing varies substantially by clinical note section. Discharge
summaries are 83.6\% unmodified while history of present illness (HPI)
sections are only 49.9\% unmodified
(Appendix~\ref{app:section-framing}). The same label therefore carries
different linguistic signals depending on the documentation context in
which it appears.

\paragraph{Summary.}
Across all three framing dimensions, labels that appear uniform encode
meaningfully different clinical situations. Temporal context may be
flattened, uncertainty may be resolved or collapsed, and intent may be
treated as more determinate than the surrounding documentation supports.
These findings do not show that the labels are wrong. Rather, they show
that EHR-based suicidality labels are operationalizations of documented
clinical evidence, and that downstream NLP systems should be evaluated
with those operationalizations in view.

\section{Discussion} 
The issues we identify are not unique to ScAN. Any EHR-based
suicidality dataset will reflect some set of choices about whose
perspective is encoded, how episodes are bounded, and how ambiguity is
resolved. The question is whether these choices are examined. Below, we trace the downstream consequences of the operationalization we have described and outline steps towards making these choices more visible. 

\subsection{Possible Downstream Effects}
If training labels subsume heterogeneous linguistic framings, classifiers trained on those labels would be expected to exhibit systematic and predictable errors. The historical-marker rate among present-SI spans (27.8\%) predicts that models would learn to associate retrospective language with active ideation, potentially overflagging historical mentions in new clinical text. The gap in unmodified rates between discharge summaries and HPI sections (83.6\% vs.\ 49.9\%) predicts that classifier performance would vary substantially across note section types, even within a single hospital stay. The collapse of unsure with negative SA labels predicts that models would be less calibrated on cases involving genuine epistemic uncertainty---precisely the cases where clinical decision support is most needed. We do not test these predictions here, but each is falsifiable through standard error analysis on classifiers trained with and without label distinctions preserved. 

\subsection{Alternative Design Choices and Recommendations}

Each dimension of the operationalization we describe suggests both an
alternative dataset design and a corresponding evaluation practice
that can surface hidden assumptions even when the dataset itself
cannot be rebuilt.

For temporal framing, labeling across encounters rather than within a
single hospital stay would allow models to represent suicidality as a
recurring process. Where cross-encounter labeling is not feasible,
reporting model performance separately by clinical note section type
would reveal whether classifiers generalize across the variation in
linguistic framing we observed---for example, between discharge
summaries and HPI notes.

For intent resolution, retaining ``unsure'' as a distinct modeling
category rather than merging it with ``negative'' would preserve
clinically meaningful uncertainty in the label space. This matters
not only for model calibration but also for clinicians who may
encounter decision-support tools trained on these labels: a system
that treats genuinely ambiguous cases as resolved negatives risks
understating uncertainty in precisely the situations where clinical
judgment is most needed. At minimum, pipelines that collapse
ambiguous categories should evaluate performance on each constituent
category before and after the merger.

For the documentation-mediated dimension, two related limitations
apply: what text is available for annotation, and how consistently
that text is interpreted. Clinical notes are authored by clinicians
and reflect their documentation practices, even when they incorporate
patient-reported information. Incorporating patient-authored
sources---such as self-report instruments, intake questionnaires, or
patient-facing communications---alongside clinician-authored notes
would allow datasets to represent suicidality across multiple
perspectives rather than through clinical documentation alone.
Independently, multi-annotator labeling with reported inter-annotator
agreement would make transparent how consistently annotators interpret
the clinical text that is available. 

Each of these alternatives introduces practical trade-offs:
longitudinal labeling requires cross-encounter linking infrastructure,
preserving ambiguous categories adds complexity to model training and
evaluation, and multi-annotator schemes increase annotation effort.
These costs are real, but they are the costs of making the
operationalization an explicit design choice rather than an invisible
default. More broadly, dataset documentation should state which
operationalization choices a pipeline inherits---how cohorts are
seeded, at what level labels are aggregated, whether ambiguous
categories are collapsed, and how many annotators contributed---so
that downstream users, including clinicians evaluating
decision-support systems, can assess what the labels represent
before acting on model outputs.

\section{Conclusion}
We have argued that EHR-based suicidality datasets should be read as
operationalizations shaped by governance, cohort design, and
annotation practice, rather than as neutral ground truth. Using ScAN
as a case study, we showed that these choices produce labels that
flatten temporality, resolve ambiguity, and encode clinician-documented
judgments. Our linguistic analysis confirmed that identical labels
subsume spans with different temporal, negation, and uncertainty
profiles. 

None of these observations are criticisms of ScAN's creators, who made
reasonable decisions under real constraints. The point is that these
decisions constitute an operationalization, and that
operationalization should be visible to downstream users. When labels
are treated as ground truth without reference to the assumptions that
produced them, the gap between what the dataset represents and what
models are trained to predict becomes difficult to assess. Making these assumptions visible is a necessary first step toward
clinical NLP systems that are transparent about what they predict.
Toward that end, we have outlined testable predictions about
downstream model behavior and practical recommendations that can help move this conversation
forward.

\section{Limitations}
 
Our case study examines one dataset from one academic medical center.
The specific distributions we report (e.g., the rate of historical
markers in present-SI spans) may
differ in datasets with other governance models, coding systems, or
annotation protocols. Our argument is structural: any pipeline that
seeds cohorts from billing codes, labels at the hospital-stay level,
and resolves intent from clinical text will encode analogous
assumptions. We do not train or evaluate classifiers on ScAN, so we
cannot say whether the within-label heterogeneity we identify
produces systematic model errors; establishing that link is a natural
next step.
 
Our linguistic analysis covers negation, temporality, and uncertainty
using lexical indicators and MedSpaCy's ConText algorithm, both of
which operate at the span level without syntactic parsing or
coreference resolution. This means we do not capture attribution,
cross-note temporal reasoning, or scope ambiguities where a modifier
may or may not govern the suicidality mention. Because our detection
methods are conservative, the heterogeneity we report is likely a
lower bound. 
 
\bibliography{custom}

@article{schwartzRacialDisparitiesPsychotic2014,
  title = {Racial Disparities in Psychotic Disorder Diagnosis: {{A}} Review of Empirical Literature},
  shorttitle = {Racial Disparities in Psychotic Disorder Diagnosis},
  author = {Schwartz, Robert C},
  year = 2014,
  journal = {World Journal of Psychiatry},
  volume = {4},
  number = {4},
  pages = {133},
  issn = {2220-3206},
  doi = {10.5498/wjp.v4.i4.133},
  urldate = {2026-05-22},
  langid = {english}
}

@article{mccabeHowHealthcareProfessionals2017,
  title = {How Do Healthcare Professionals Interview Patients to Assess Suicide Risk?},
  author = {McCabe, Rose and Sterno, Imren and Priebe, Stefan and Barnes, Rebecca and Byng, Richard},
  year = 2017,
  month = dec,
  journal = {BMC Psychiatry},
  volume = {17},
  number = {1},
  pages = {122},
  issn = {1471-244X},
  doi = {10.1186/s12888-017-1212-7},
  urldate = {2026-05-22},
  langid = {english}
}

@article{roushMentalHealthProfessionals2018,
  title = {Mental {{Health Professionals}}' {{Suicide Risk Assessment}} and {{Management Practices}}: {{The Impact}} of {{Fear}} of {{Suicide-Related Outcomes}} and {{Comfort Working With Suicidal Individuals}}},
  shorttitle = {Mental {{Health Professionals}}' {{Suicide Risk Assessment}} and {{Management Practices}}},
  author = {Roush, Jared F. and Brown, Sarah L. and Jahn, Danielle R. and Mitchell, Sean M. and Taylor, Nathanael J. and Quinnett, Paul and Ries, Richard},
  year = 2018,
  month = jan,
  journal = {Crisis},
  volume = {39},
  number = {1},
  pages = {55--64},
  issn = {0227-5910, 2151-2396},
  doi = {10.1027/0227-5910/a000478},
  urldate = {2026-05-22},
  langid = {english}
}

@article{Anderson_2015,
  title={Monitoring suicidal patients in primary care using electronic health records},
  author={Anderson, Heather D and Pace, Wilson D and Brandt, Elias and Nielsen, Rodney D and Allen, Richard R and Libby, Anne M and West, David R and Valuck, Robert J},
  journal={The Journal of the American Board of Family Medicine},
  volume={28},
  number={1},
  pages={65--71},
  year={2015},
  publisher={American Board of Family Medicine}
}

@article{belsher2019prediction,
  title={Prediction models for suicide attempts and deaths: a systematic review and simulation},
  author={Belsher, Bradley E and Smolenski, Derek J and Pruitt, Larry D and Bush, Nigel E and Beech, Erin H and Workman, Don E and Morgan, Rebecca L and Evatt, Daniel P and Tucker, Jennifer and Skopp, Nancy A},
  journal={JAMA psychiatry},
  volume={76},
  number={6},
  pages={642--651},
  year={2019}
}

@inproceedings{chancellor2019,
  title={A taxonomy of ethical tensions in inferring mental health states from social media},
  author={Chancellor, Stevie and Birnbaum, Michael L and Caine, Eric D and Silenzio, Vincent MB and De Choudhury, Munmun},
  booktitle={Proceedings of the conference on fairness, accountability, and transparency},
  pages={79--88},
  year={2019}
}

@inproceedings{ernala2019methodological,
  title={Methodological gaps in predicting mental health states from social media: Triangulating diagnostic signals},
  author={Ernala, Sindhu Kiranmai and Birnbaum, Michael L and Candan, Kristin A and Rizvi, Asra F and Sterling, William A and Kane, John M and De Choudhury, Munmun},
  booktitle={Proceedings of the 2019 chi conference on human factors in computing systems},
  pages={1--16},
  year={2019}
}

@article{franklin2017risk,
  title={Risk Factors for Suicidal Thoughts and Behaviors: 
         A Meta-Analysis of 50 Years of Research},
  author={Franklin, Joseph C. and Ribeiro, Jessica D. and Fox, Kathryn R. and 
          Bentley, Kate H. and Kleiman, Evan M. and Huang, Xieyining and 
          Musacchio, Katherine M. and Jaroszewski, Adam C. and 
          Chang, Bernard P. and Nock, Matthew K.},
  journal={Psychological Bulletin},
  volume={143},
  number={2},
  pages={187--232},
  year={2017},
  doi={10.1037/bul0000084}
}

@article{harkema2009context,
  title={ConText: an algorithm for determining negation, experiencer, and temporal status from clinical reports},
  author={Harkema, Henk and Dowling, John N and Thornblade, Tyler and Chapman, Wendy W},
  journal={Journal of biomedical informatics},
  volume={42},
  number={5},
  pages={839--851},
  year={2009},
  publisher={Elsevier}
}

@inproceedings{harrigian2020state,
  title={On the state of social media data for mental health research},
  author={Harrigian, Keith and Aguirre, Carlos and Dredze, Mark},
  booktitle={Proceedings of the Seventh Workshop on Computational Linguistics and Clinical Psychology: Improving Access},
  pages={15--24},
  year={2021}
}

@article{Johnson_2016,
  title={MIMIC-III clinical database (version 1.4)},
  author={Johnson, Alistair and Pollard, Tom and Mark, Roger},
  journal={PhysioNet},
  volume={10},
  number={C2XW26},
  pages={2},
  year={2016}
}

@article{Kessler_2020,
  title={Suicide prediction models: a critical review of recent research with recommendations for the way forward},
  author={Kessler, Ronald C and Bossarte, Robert M and Luedtke, Alex and Zaslavsky, Alan M and Zubizarreta, Jose R},
  journal={Molecular psychiatry},
  volume={25},
  number={1},
  pages={168--179},
  year={2020},
  publisher={Nature Publishing Group UK London}
}

@article{large2016metaanalysis,
  title={Meta-Analysis of Longitudinal Cohort Studies of Suicide Risk 
         Assessment among Psychiatric Patients: Heterogeneity in Results 
         and Lack of Improvement over Time},
  author={Large, Matthew and Kaneson, Muthusamy and Myles, Nicholas and 
          Myles, Hannah and Gunaratne, Pramudie and Ryan, Christopher},
  journal={PLOS ONE},
  volume={11},
  number={6},
  pages={e0156322},
  year={2016},
  doi={10.1371/journal.pone.0156322}
}

@inproceedings{Rawat_2022,
  title={ScAN: suicide attempt and ideation events dataset},
  author={Rawat, Bhanu Pratap Singh and Kovaly, Samuel and Yu, Hong and Pigeon, Wilfred},
  booktitle={Proceedings of the 2022 Conference of the North American Chapter of the Association for Computational Linguistics: Human Language Technologies},
  pages={1029--1040},
  year={2022}
}

@article{sun2022negative,
  title={Negative Patient Descriptors: Documenting Racial Bias In The Electronic Health Record: Study examines racial bias in the patient descriptors used in the electronic health record.},
  author={Sun, Michael and Oliwa, Tomasz and Peek, Monica E and Tung, Elizabeth L},
  journal={Health Affairs},
  volume={41},
  number={2},
  pages={203--211},
  year={2022}
}
\newpage 
\appendix
 
\section{Framing Analysis: Definitions}
\label{app:framing-method}
 
\paragraph{Lexical keyword lists.}
The following case-insensitive terms are matched within each
three-sentence context window:
 
\begin{itemize}
    \item \textbf{Negation:} \emph{no, not, denies, denied, without,
          never, none}
    \item \textbf{Historical:} \emph{history of, h/o, hx of, previous,
          prior, past, ago, years ago, months ago}
    \item \textbf{Uncertainty:} \emph{may, might, possibly, unclear,
          uncertain, equivocal, question of, rule out, r/o}
\end{itemize}
 
\noindent Lexical matching does not resolve scope: a keyword may modify
a concept unrelated to suicidality. We therefore run MedSpaCy's
ConText algorithm \citep{harkema2009context} in parallel with default
\texttt{ConTextComponent} settings, retaining only modifiers whose scope
overlaps with the annotated span.

 \paragraph{Composite measure.}
A span is framed as \emph{unmodified} when both methods agree on the absence
of negation, historical, and uncertainty modifiers. This conservative
definition means some modified spans may carry modifiers that do not
pertain to the suicidality judgment.
\section{Signal Agreement Rates}
\label{app:signal-agreement}
 
Table~\ref{tab:signal-agreement} reports span-level agreement between
the lexical and ConText detectors ($n = 15{,}585$).
 
\begin{table}[h!]
\centering
\small
\caption{Base rates and agreement between lexical and ConText
detectors. The uncertainty divergence ($9.4{\times}$) reflects how
often terms like \emph{may} modify non-suicidality concepts.}
\label{tab:signal-agreement}
\begin{tabular}{@{}lrrrr@{}}
\hline
\textbf{Signal} & \textbf{Lex} & \textbf{Ctx} & \textbf{Agree} & \textbf{Ratio} \\
\hline
Negation     & 6.3\%  & 7.6\%  & 92.1\% & $0.8{\times}$ \\
Historical   & 19.6\% & 8.4\%  & 86.6\% & $2.3{\times}$ \\
Uncertainty  & 9.4\%  & 1.0\%  & 91.2\% & $9.4{\times}$ \\
\hline
\end{tabular}
\end{table}
 
\section{Category-Level Framing}
\label{app:category-detail}
 
Tables~\ref{tab:framing-sa-full} and~\ref{tab:framing-si-full} extend
the summaries in Section~\ref{sec:framing} to all annotation categories.
Lex-H / Ctx-H = lexical / ConText historical; Lex-N / Ctx-N = negation;
Lex-U = lexical uncertainty (ConText uncertainty ${\leq}\,1\%$
everywhere, omitted).

\begin{table*}[h!]
\centering
\small
\caption{Framing profiles for SA categories ($n=14{,}048$), ordered
by \% Unmodified.}
\label{tab:framing-sa-full}
\begin{tabular}{@{}llrrrrrrr@{}}
\hline
\textbf{Cat.} & \textbf{Description} & \textbf{$n$} &
\textbf{\% Unmod.} &
\textbf{Lex-H} & \textbf{Ctx-H} &
\textbf{Lex-N} & \textbf{Ctx-N} &
\textbf{Lex-U} \\
\hline
T71       & Asphyxiation               &     440 & 86.6 &  9.6 &  2.1 &  0.5 &  0.7 &  3.6 \\
X71--X83  & Intentional self-harm      & 1{,}271 & 83.3 & 12.1 &  5.0 &  1.1 &  1.6 &  1.7 \\
T51--T65  & Toxic subst.\ (non-pharma) &     531 & 74.0 & 16.2 &  6.8 &  2.8 &  5.1 &  4.7 \\
T36--T50  & Pharma.\ poisoning         & 8{,}635 & 69.5 & 16.3 &  6.5 &  2.5 &  4.7 & 10.1 \\
Unsure    & Annotator unsure           & 2{,}237 & 61.6 & 19.3 &  2.3 &  6.8 &  6.2 & 17.1 \\
T14.91    & Unspecified intent         &     770 & 19.0 & 67.8 & 50.4 &  3.4 &  6.6 &  5.1 \\
N/A       & No ICD code                &     164 & 17.1 & 37.8 & 23.8 & 50.0 & 56.1 &  9.2 \\
\hline
\end{tabular}%
\end{table*}
 
\begin{table*}[h!]
\centering
\small
\caption{Framing profiles for SI spans ($n=1{,}537$) by status label.}
\label{tab:framing-si-full}
\begin{tabular}{@{}lrrrrrrr@{}}
\hline
\textbf{Status} & \textbf{$n$} &
\textbf{\% Unmod.} &
\textbf{Lex-H} & \textbf{Ctx-H} &
\textbf{Lex-N} & \textbf{Ctx-N} &
\textbf{Lex-U} \\
\hline
Present & 897 & 52.2 & 27.8 & 12.5 &  7.0 & 11.8 & 7.1 \\
Absent  & 554 & 13.5 & 16.1 &  9.2 & 65.2 & 57.2 & 3.1 \\
N/A     &  86 & 23.3 & 14.0 &  8.1 & 52.3 & 37.2 & 11.6 \\
\hline
\end{tabular}
\end{table*}
 
\section{Section-Level Framing}
\label{app:section-framing}
 
Table~\ref{tab:section-framing-full} reports framing rates by clinical
note section. Sections with $n < 40$ are included for completeness.
\% SI = share of SI spans in section.
 
\begin{table*}[h!]
\centering
\small
\caption{Framing by clinical note section, ordered by \% Unmodified.}
\label{tab:section-framing-full}
\begin{tabular}{@{}lrrrrrrr@{}}
\hline
\textbf{Section} & \textbf{$n$} &
\textbf{\% Unmod.} &
\textbf{Lex-H} & \textbf{Ctx-H} &
\textbf{Lex-N} & \textbf{Ctx-N} &
\textbf{\% SI} \\
\hline
Discharge summary   & 3{,}358 & 83.6 &  9.3 &  2.4 &  1.5 &  1.7 &  4.0 \\
Assessment / plan   & 4{,}262 & 66.1 & 16.5 &  6.0 &  6.3 &  7.7 & 12.4 \\
Medical history     &     743 & 61.5 & 28.1 & 15.3 &  3.4 &  4.7 & 10.5 \\
Other               &      31 & 61.3 & 32.3 &  6.5 &  3.2 &  3.2 &  6.5 \\
Chief complaint     & 2{,}567 & 56.7 & 23.3 &  9.5 &  6.3 & 12.5 &  7.8 \\
Impression          &     840 & 55.4 & 21.6 & 11.6 &  8.6 &  6.8 & 10.2 \\
Social history      & 2{,}270 & 51.7 & 23.6 & 12.3 &  9.1 & 11.2 & 12.5 \\
HPI                 & 1{,}474 & 49.9 & 33.4 & 16.2 & 12.8 &  9.2 & 14.9 \\
Psychiatric history &      40 & 45.0 & 35.0 & 10.0 & 12.5 &  2.5 & 15.0 \\
\hline
\end{tabular}%
\end{table*}
 
\section{Demographic Analysis}
\label{app:demographics}
 
Analyses use 486 admissions with structured data from MIMIC-III's
\texttt{ADMISSIONS} and \texttt{PATIENTS} tables joined to ScAN
annotations. We report Pearson's $\chi^2$ tests with uncorrected
$p$-values; given small subgroups and the exploratory scope, we do not
apply multiple-comparison corrections.
 
\paragraph{Subgroup sizes.}
Insurance: Private 181, Medicaid 122, Medicare 103, Government 56,
Self Pay 22, missing 2. Gender: Female 257, Male 227, missing 2.
Race/ethnicity (collapsed): White 339,
Unknown/Not Specified\footnote{Includes \textsc{unknown/not specified}
($n{=}54$), \textsc{unable to obtain} ($n{=}17$), \textsc{patient
declined} ($n{=}3$), missing ($n{=}2$).} 76, Black/African American 22,
Hispanic/Latino 18, Other 22, Asian 9. No non-White group exceeds 22
admissions; race/ethnicity comparisons are not reported.

Tables \ref{tab:gender}, \ref{tab:age}, \ref{tab:ethnicity}, \ref{tab:language}, \ref{tab:insurance}, \ref{tab:marital} and \ref{tab:religion} show the respective distributions of gender, age, ethnicity, language spoken, insurance status, marital status, and religion of patients in the dataset. Additionally, the distribution of living vs. deceased patients is reported in Table \ref{tab:mortality}.

\paragraph{Ambiguities.} Ambiguities were noted in the gender, language, and ethnicity data. ScAN included one male-to-female transgender individual. While the MIMIC-III documentation defines gender as "the genotypical sex of the patient" \cite{Johnson_2016}, the transgender patient's gender was listed as Female, which is not their genotypical sex. Further, MIMIC-III, and by extension ScAN, uses non-standard 4 letter language codes. While some codes are easy to infer, such as \emph{SPAN} for Spanish, other codes such as \emph{CAPE} and \emph{PTUN} are not intuitive and have no documented definition. Finally, ethnicity data is provided with varying levels of specificity. For example, an individual could be listed as \emph{White}, \emph{White -- Brazilian}, \emph{White -- Russian}, or \emph{White -- Other European}. However, we acknowledge that it may not always be possible to obtain a patient's specific ethnicity.

\begin{table*}[h!]
\centering
\small
\caption{Demographic association tests at or near significance. All
other comparisons returned $p > 0.10$.}
\label{tab:demo-tests}
\begin{tabular}{@{}llrl@{}}
\hline
\textbf{Comparison} & \textbf{Outcome} & \textbf{$p$} & \\
\hline
Medicaid vs.\ Private & ${\geq}1$ unsure SA & 0.014 & * \\
\quad \small{38/122 (31.1\%) vs.\ 33/181 (18.2\%)} & & & \\
\hline
Female vs.\ Male & ${\geq}1$ historical & 0.087 & $\dagger$ \\
\quad \small{221/257 (86.0\%) vs.\ 181/227 (79.7\%)} & & & \\
\hline
Female vs.\ Male & ${\geq}1$ uncertainty & 0.056 & $\dagger$ \\
\quad \small{159/257 (61.9\%) vs.\ 120/227 (52.9\%)} & & & \\
\hline
\multicolumn{4}{@{}l}{\small * $p < 0.05$; \; $\dagger$ $p < 0.10$ (marginal)} \\
\end{tabular}
\end{table*}

\begin{table}[ht]
\small
\centering
\begin{tabular}{lr}
\hline
\textbf{Gender} & \textbf{\%} \\
\hline
Female & 50.9 \\
Male   & 49.1 \\
\hline
\end{tabular}
\caption{Gender distribution of ScAN patients. }
\label{tab:gender}
\end{table}

\begin{table}[ht]
\small
\centering
\begin{tabular}{lr}
\hline
\textbf{Age Group} & \textbf{\%} \\
\hline
18--24 & 13.5 \\
25--34 & 21.6 \\
35--44 & 22.8 \\
45--54 & 25.5 \\
55--64 &  9.7 \\
65--74 &  5.1 \\
75+    &  1.9 \\
\hline
\end{tabular}
\caption{Age distribution of ScAN patients.}
\label{tab:age}
\end{table}

\begin{table}[ht]
\small
\centering
\begin{tabular}{lr}
\hline
\textbf{Ethnicity} & \textbf{\%} \\
\hline
White                              & 70.0 \\
Unknown / Other / Declined         & 16.4 \\
Black / African American           &  5.9 \\
Hispanic or Latino                 &  3.6 \\
Asian                              &  1.7 \\
European                           &  0.7 \\
Black / Cape Verdean               &  0.6 \\
Multi-race                         &  0.4 \\
South American                     &  0.3 \\
American Indian / Alaska Native    &  0.1 \\
Native Hawaiian / Pacific Islander &  0.1 \\
\hline
\end{tabular}
\caption{Ethnicity distribution of ScAN patients. 'European' includes the categories 'Portugese', 'White -- Russian', and 'White -- Other European'. 'Asian' includes the categories 'Asian', 'Asian -- Japanese', 'Asian -- Chinese', 'Asian -- Vietnamese'. South American includes the categories 'South American' and 'White -- Brazilian'. 'Hispanic or Latino' includes the categories 'Hispanic or Latino', 'Hispanic/Latino -- Puerto Rican', and 'Hispanic/Latino -- Dominican'. }
\label{tab:ethnicity}
\end{table}

\begin{table}[ht]
\small
\centering
\begin{tabular}{lr}
\hline
\textbf{Language} & \textbf{\%} \\
\hline
English (ENGL)   & 92.3 \\
Spanish (SPAN)   &  2.4 \\
Unknown (PTUN)   &  1.8 \\
Russian (RUSS)   &  1.0 \\
Cantonese (CANT) &  0.4 \\
Portugese (PORT) &  0.4 \\
Unknown (CAPE)   &  0.4 \\
Other            &  1.3 \\
\hline
\end{tabular}
\caption{Language distribution of ScAN patients. 'Other' includes Japanese (JAPA), Arabic (ARAB), Mandarin (MAND), American Sign Language (AMER), French (FREN), and Haitian (HAIT). Each of these languages have 0.2\% speakers in ScAN.}
\label{tab:language}
\end{table}

\begin{table}[ht]
\small
\centering
\begin{tabular}{lr}
\hline
\textbf{Insurance Type} & \textbf{\%} \\
\hline
Private    & 34.9 \\
Medicaid   & 26.2 \\
Medicare   & 24.1 \\
Government & 10.8 \\
Self-pay   &  4.0 \\
\hline
\end{tabular}
\caption{Insurance distribution of ScAN patients.}
\label{tab:insurance}
\end{table}

\begin{table}[ht]
\small
\centering
\begin{tabular}{lr}
\hline
\textbf{Marital Status} & \textbf{\%} \\
\hline
Single    & 62.1 \\
Married   & 25.3 \\
Divorced  &  7.6 \\
Widowed   &  2.8 \\
Separated &  1.2 \\
Unknown   &  1.0 \\
\hline
\end{tabular}
\caption{Marital status distribution of ScAN patients.}
\label{tab:marital}
\end{table}

\begin{table}[ht]
\small
\centering
\begin{tabular}{lr}
\hline
\textbf{Religion} & \textbf{\%} \\
\hline
Catholic            & 57.1 \\
Other               & 13.7 \\
Protestant / Quaker & 12.7 \\
Jewish              & 11.4 \\
Christian -- Other  &  3.8 \\
Buddhist            &  1.3 \\
\hline
\end{tabular}
\caption{Religion distribution of ScAN patients among the 45.4\% who elected to provide their information. The remaining 54.6\% either did not specify their religion or this information was unobtainable.}
\label{tab:religion}
\end{table}

\begin{table}[ht]
\small
\centering
\begin{tabular}{lr}
\hline
\textbf{Mortality} & \textbf{\%} \\
\hline
Alive                     & 84.9 \\
Deceased (post-discharge) & 10.2 \\
Deceased (during visit)   &  4.9 \\
\hline
\end{tabular}
\caption{Mortality distribution of ScAN patients.}
\label{tab:mortality}
\end{table}
 
\end{document}